\documentclass[12pt, titlepage]{article}
\usepackage{fullpage}
\usepackage{fullname}
\usepackage{graphicx}

\long\def\comment#1{{}}

\title{A Memory-Based Approach to Learning Shallow Natural Language Patterns}

\author{Shlomo Argamon-Engelson \and Ido Dagan \and Yuval Krymolowski \\
        Department of Mathematics and Computer Science\\
	Bar-Ilan University\\
	52900 Ramat Gan, Israel\\
 {\tt \{argamon,dagan,yuvalk\}@cs.biu.ac.il}
}

\def\th{\theta}

\begin{document}

\maketitle

\begin{abstract}
Recognizing shallow linguistic patterns, such as basic syntactic
relationships between words, is a common task in applied natural
language and text processing. The common practice for approaching this
task is by tedious manual definition of possible pattern structures,
often in the form of regular expressions or finite automata. This
paper presents a novel memory-based learning method that recognizes
shallow patterns in new text based on a bracketed training corpus. The
examples are stored as-is, in efficient data structures.
Generalization is performed on-line at recognition time
by comparing subsequences of the new text to positive and negative
evidence in the corpus. This way, no information in the training is
lost, as can happen in other learning systems that construct a single
generalized model at the time of training. The paper presents
experimental results for recognizing noun phrase, subject-verb and
verb-object patterns in English.
\end{abstract}



\setcounter{page}{1}
\pagestyle{plain}

\section{Introduction}

Identifying local patterns of syntactic sequences and relationships is a
fundamental task in natural language processing (NLP). Such patterns
may correspond to syntactic phrases, like noun phrases, or to pairs
of words that participate in a syntactic
relationship, like the  heads of a verb-object relation. Such
patterns have been found useful in various
application areas, including information extraction, text
summarization, and bilingual alignment. Syntactic patterns are  useful
also for many basic computational linguistic
tasks, such as statistical word similarity and various disambiguation
problems.

One approach for detecting syntactic patterns is to obtain a full
parse of a sentence and then extract the required patterns. However,
obtaining a complete parse tree for a sentence is difficult in many
cases, and may not be necessary at all for identifying most instances
of local syntactic patterns.

An alternative approach is to avoid the complexity of full parsing and instead
analyse a sentence at the level of phrases and the relations between them,
this is the task of {\em {shallow parsing}} \cite{Abney91,gref:acl93}.
In contrast to full parsing, shallow parsing can be achieved using
{\em {local}} information. Previous works (cf.~section~\ref{sec:lrn-shallow}) 
have shown that it is possible to identify most instances of
shallow syntactic patterns by rules that examine only the pattern itself
and its nearby context. Often, the rules are applied to sentences that
were tagged by part-of-speech (POS) and are phrased by some form of
regular expressions or finite state automata. 

Manual writing of local syntactic rules has become a common practice
for many applications. However, writing rules is
often tedious and time consuming.  Furthermore, extending the
rules to different languages or sub-language domains can require
substantial resources and expertise that are often not available. As in many
areas of NLP, a {\em learning} approach is appealing.

Abney \shortcite{Abney91} introduced the notion of {\em chunks}, denoting
sequences
of words with a certain syntactic function (more in section~\ref{sec:bkg:sp}).
The task of dividing a sentence into meaningful sequences of words is thus
referred to as {\em chunking}.
The most studied chunking task is that of identifying noun phrases (NPs,
e.g.~\namecite{Church88}, \namecite{ram-mar-95}, \namecite{cardie-pierce-98},
\namecite{veenstra-np-98}),
naturally due to their central role in a sentence.

This paper presents a novel general learning approach for recognizing local
sequential patterns, that falls within
the  memory-based learning paradigm. The method learns from a POS
tagged training corpus in which all instances of the target pattern are marked
(bracketed). Subsequences of the training examples are stored as-is in a trie,
thereby facilitating a linear-time search for subsequences in the corpus.
While the presented algorithm is oriented at recognizing sequences, it is
useful also for grammatical {\em relations} such as subject-verb, and verb-object.
We present these applications, in which the relations are represented as
sequences encompassing the relevant words. The algorithm is therefore suitable
for learning to perform tasks involved in shallow parsing.

The memory-based nature of the presented algorithm
stems from its induction strategy: a sequence is recognized as an instance of
the target pattern by examining the {\em raw} training
corpus, searching for relevant positive and negative evidence.
No model is created for storing the training corpus, and the raw data are not
converted to any other representation.
Modelling lies in the choice of the kinds of data retrieved from the stored
corpus, that is, in the substrings being searched in the memory.
We believe this choice to be highly relevant for linguistic data.

The POS tag set used throughout the paper is the Penn TreeBank set
\cite{TB_corpus}:
{\tt DT} = determiner, {\tt ADJ} = adjective, {\tt RB} = adverb,
{\tt VB}=verb, {\tt PP}=preposition,
{\tt NN} = singular noun, and {\tt NNP} = plural noun.
As an illustrating example, Suppose we want to decide whether the candidate
sequence
\begin{verbatim}
  DT ADJ ADJ NN NNP
\end{verbatim}
is an NP using information from the training corpus. Finding
the entire sequence as-is several times in the corpus
would yield an exact match. Due to data sparseness, however, that cannot
always be expected.

A somewhat weaker match may be obtained if we consider {\em tiles},
sub-parts of the candidate sequence. For example, suppose the corpus
contains noun phrases with the following structures:
\begin{verbatim}
   (1) DT ADJ ADJ NN NN
   (2) DT ADJ NN NNP
\end{verbatim}
The first structure provides positive evidence that the
sequence
`\texttt {DT~ADJ~ADJ~NN}' is a possible NP prefix while the
second structure provides evidence for
`\texttt {ADJ~NN~NNP}' being an NP suffix. Together, these
two training instances provide positive evidence that {\em covers} the
entire candidate. Considering evidence for sub-parts of the pattern
enables us to generalize over the exact structures present in
the corpus. Similarly, the algorithm considers the negative evidence for
such sub-parts by noting where they occur in the corpus without being
a corresponding part of a target-pattern instance. Surrounding context and
evidence overlap are considered as well.

Other implementations of the memory-based paradigm for NLP tasks include
\namecite{daelemans-mb-96}, for POS tagging; \namecite{cardie93}, for
syntactic and semantic tagging; and \namecite{Stanfill-Waltz86},
for word pronunciation.
  In all these works, examples are represented as sets of features and the
  induction is carried out by finding the most similar cases.
The memory-based works of \namecite[DOP]{Bod92CO}
for parsing, and \namecite{yvon-96} for pronunciation use the {\em raw}
form of the data, rather than encode it as features.
The method presented here is similar in that it makes use of {\em raw
sequential} data, and generalizes by reconstructing test examples
from different pieces of the training data.

Previous related work is described in section~\ref{sec:bkg}.
Section~\ref{sec:alg} describes the inference algorithm formally;
section \ref{sec:eval} presents experimental results for three
target syntactic patterns in English, along with comparison with related
results. 

\section{Background}
\label{sec:bkg}

We present here a brief overview of the state-of-the-art in shallow
parsing, covering both hand-crafted and learnable parsers.  In
addition, since shallow parsing can be viewed as a sub-task of full
parsing, we also discuss relevant methods for learning full parsing.
Section~\ref{sec:bkg:sp} presents some of the current hand-crafted
shallow parsers, while sections~\ref{sec:lrn-full}
and~\ref{sec:lrn-shallow} present methods for learning full and
shallow parsing respectively.

\subsection{Shallow Parsers with Hand-Written Rules}
\label{sec:bkg:sp}

Much work to date on shallow parsing has been based on hand-crafted
sets of rules, generally using a probabilistic model to choose
between parse alternatives.

\namecite[1996]{Abney91} has pioneered work on shallow parsing.
His work was motivated by psycholinguistic evidence
\cite{gee-grosjean}, indicating that language processing involves
dividing the sentence into {\em performance structures} ---
corresponding to pauses and intonation changes during speech.  Abney
introduced the concept of {\em chunks}, defined as consisting of `a
single content word surrounded by a constellation of function words,
matching a fixed template'. His chunk parser operates in two phases:
a {\em chunker} which offers potential chunks, and an {\em attacher}
which resolves attachment ambiguities and selects the final
chunks. The chunker makes use of POS data, whereas the attacher
requires lexical information. That way, lexical information is used
only for the tasks for which it is more important, and simpler POS
information is used for the more basic task of chunking.  Both parts
of Abney's chunk parser are implemented as non-deterministic LR
parsers. The distinction between chunking and attachment is common to
all the systems presented in this section.

\namecite{ait-mokhtar-chanod-97} presented a sequence of finite-state
transducers for extracting subjects and objects from POS-tagged French
texts.  Extraction is incremental, each processing phase is carried
out by a dedicated transducer, which prepares the input for the next
phase. The system relies only on POS information, that is, it does not
require lexical information. For various corpora, the recall and
precision for subjects were above 90.5\% and 86.5\% respectively,
whereas for objects these figures exceeded 84.4\% and 79.9\% .

Much shallow parsing effort has been motivated by
information-extraction (IE) tasks such as MUC. For example,
the FASTUS \cite{Appelt:93IJ} system uses cascaded, non-deterministic
finite-state automata for extracting noun groups, verb groups, and
particles. As an IE system, FASTUS is built for cases where only part
of the text is relevant, and the target patterns can be represented in
a simple rigid fashion.

\namecite{schiller-96} presented a finite-state multilingual system for
NP detection. This system is also intended for simple phrases, without
coordinations or relative clauses.

The SPARKLE (Shallow PARsing for acquisition of Knowledge for Language
Engineering, {\tt http://www.ilc.pi.cnr.it/sparkle/sparkle.htm}) project
aims mainly in developing
`robust and portable tools leading to commercial applications devoted
to the management of multilingual information in electronic
form'. Phrasal-level syntactic analysis is essential for
the objective of the project; the resulting parses will serve for acquiring
lexical information. Part of the SPARKLE project is thus devoted to
developing shallow parsers.

\begin{itemize}
\item The shallow parser for English is developed in Cambridge and Sussex
universities. Parsing is carried out by a generalized LR parser, which
uses a unification-based phrasal grammar of POS tags. The parser
performs disambiguation based on a probabilistic model, that is, the
rules are fixed, but their probabilities are being learned. The
reported recall and precision are 82.5\%/83\% for phrases (without
lexical information) and 88.1\%/88.2\% for grammatical relations
(including lexical information).

\item In the system for German, developed at the University of Stuttgart,
the grammar rules are written in a bottom-up fashion. The parser is a
standard chart parser, extended to include head-markings for
lexicalization. The parser is also enhanced to carry out EM estimation
of hand-written CFG rule probabilities. That way, both parsing and
lexical acquisition are integrated in a single process. Final
evaluation data were not available.

\item The system developed for French, at Rank-Xerox Research Centre (RXRC)
is a collection of finite-state transducers. The input is a POS-tagged
text, and each transducer inserts markup symbols for the corresponding
pattern. Parsing is carried out by invoking the transducers bottom-up,
starting from the easier tasks, using the transducer output as an
input to the next one.  Once the text is marked, another transducer
identifies the head words. That information is used in the last phase,
by a transducer which identities syntactic functions.
\end{itemize}

\subsection{Learning: Full Parses}
\label{sec:lrn-full}

Full parsing learning methods are applicable in general to shallow parsing,
where they can be used for extracting partial parses.

Some methods aim at estimating probabilities of hand-written grammar
rules, based on annotated corpora. These include the works of
\namecite{per-schabes-92} for CFG, \namecite{chit-grish-90}
 for context-sensitive grammar,
\namecite{eisner-96} for dependency parsing,
\namecite{mag-mar-91} for bottom-up chart parsing, and \namecite{bris-car-93}
for attribute-value grammars.  In the rest of the section
we present works which do not use a certain set of grammar rules, but
induce the parsing from statistical features extracted from the training data.

\namecite{Brill93} used transformation-based-learning for learning to 
produce an unlabeled parse tree. Given a POS-tagged sentence, the learned
transformations manipulated insertion and deletion of left and right
parenthesis
depending on the current POS tag and, possibly, the POS tag of the
previous or next word.

SPATTER \cite{magerman95}, is a learning algorithm for parsing which
makes use only of hierarchical structure information. It learns a
decision-tree model for tagging and parsing, where parsing decisions build a
hierarchical structure in a bottom-up manner. The questions at the nodes
of the tree take into account neighboring as well as child nodes; the
result is a complete parse tree. It scores 84\%
recall and 84.3\% precision on Penn TreeBank WSJ data.

The parser of  \namecite{collins-97} is based on statistics of lexical
dependencies, subcategorization and wh-movement. His parser
scored 88.1\% recall and 87.5\% precision on Penn TreeBank WSJ data,
currently the best result on that dataset.

Recently, \namecite{sekine-phd} built a system based on learning parse rules
for five non-terminals: sentence, subordinate sentence, infinitive
sentence, NP, and base NP; the system also uses lexical dependency
information. In addition his system is capable of performing `fitted
parsing' --- combining partial trees or chunks built by these rules,
producing a complete parse. With combined chunks restricted to {\tt S}
nodes or punctuation marks, less than 1\% of the test sentences
required fitting. Even after 40,000 training sentences, each new
sentence produced, on the average, a new rule. The parsing rules
were mapped to an automaton in order to handle their large number,
best-first search and Viterbi search were used for optimization.

The maximum-entropy parser by \namecite{rat-97}, operates in three
passes: tagging, chunking, and building hierarchical structure. A
maximum-entropy model is employed at each phase. The features which
the models test are `contextual predicates' - word, POS or chunk-tag
of neighboring (up to a distance of 2) words, or features which
represent hierarchical dependency.  Each pass prepares information for
the next one. The performance of the system on the data on which
SPATTER was tested is 86.3\% recall and 87.5\% precision.

Data Oriented Parsing (DOP), presented by \namecite{Bod92CO}, is
another example of an algorithm based purely on hierarchical structure
information. It relies on the idea of producing a parse-tree by
combining sub-trees from a memory of previous parses. For example,
suppose the training data contained the parsed sentences:
\begin{verbatim}
 (S (NP (DT The) (JJ pretty) (NN bird)) (VP (VB sings)))
 (S (NP (DT The) (NN airplane)) (VP (VB flies)))
\end{verbatim}
The system memory would contain all the sub-trees of these parse
trees; some of them are:
\begin{verbatim}
 (S (NP) (VP))
 (VP (VB))
 (NP (DT) (NN))
 (NP (DT) (JJ) (NN))
 (DT the)
 (NN bird)
 (VB flies)
\end{verbatim}
Given the new sentence: `The bird flies', the parse tree
\begin{verbatim}
 (S (NP (DT the) (NN bird)) (VP (VB flies)))
\end{verbatim}
can now be built by combining the sub-trees from the memory.

In the general case, each sub-tree is scored according to its
frequency relative to other sub-trees with the same root. Then, the
various parse alternatives are scored according to the scores of their
building blocks. Finding the best parse is NP-hard \cite{simaan-96},
therefore the DOP parsing algorithm uses a Monte-Carlo approximation and the
resulting parse is an estimation of the best one.

Note that DOP does not rely on a given set of parsing rules (nor does
it build such a set); instead, it reads the parsing trees as {\rm raw}
data and uses them as building blocks.  The algorithm presented in
this paper is similar to DOP in that it tries to build evidence for
bracketing an input sequence based on matching subsequences of the
target pattern stored in the memory.

\subsection{Learning: Shallow Parses}
\label{sec:lrn-shallow}

A number of systems have been developed for learning to perform shallow parsing.
Most of these systems learn to identify chunks such as noun phrases,
while some others learn to identify relationships between chunks (or
representative words thereof).

\namecite{Church88} uses a simple model for finding base (non-recursive)
NPs in a sequence of POS tags.  Viewing the task as a bracketing
problem, he calculated the probability of inserting open/close
brackets between POS tags. A sequence of POS tags, representing a
sentence, may be chunked into base NPs in various ways. Each chunking
alternative is scored using the probabilities, the best alternative is
then chosen. The algorithm was tested on the Brown corpus; while exact
results are not reported, they are described as encouraging. In
particular, that motivates using POS-tag information alone for NP
detection. That is important because the few dozens of POS tags
require much less resources than any collection of lexical
information.

\namecite[hereafter RM95]{ram-mar-95} viewed the problem as classification
of words. Each word was assigned a chunk-tag: `I' or `O' for words
inside or outside a chunk, and `B' for words which stand at the
beginning of a base NP that immediately follows another base NP. Thus,
in contrast to \cite{Church88}, where the sentence was chunked based
on a global criterion (the chunking with the higher score), here the
decision is {\em local}.  RM95 used transformation-based learning
\cite{Brill92}, with rule-templates referring to neighboring
words, POS tags, and chunk tags (up to a distance of 3 for words or
POS tags, and 2 for chunk tags).  Their work is the first one to
present large-scale testing of base NP learning. Training on 950K
words from the Penn TreeBank WSJ data tagged by Brill's POS tagger
\cite{Brill92}, with a test set of 50K words, they achieved a recall
of 93.5\% and precision of 93.1\% . RM95 demonstrate that the
contribution of the lexical information is about 1\%, the main
information sources were thus the neighboring POS and chunk tags.
They also report a chunk-tagging accuracy of 97.8\%.

\namecite{veenstra-np-98} recently presented an application of the IGTree
\cite{igtree} memory-based learning algorithm for NP chunking. Using the data
of RM95, POS tagged by Memory-Based Tagger \cite{daelemans-mb-96}, he also
assigned a chunk tag (`I', `O', `B') to each word.
The features included lexical
and POS information at a distance of up to two words. Some variants
have generally yielded higher recall and lower precision than RM95,
see table~\ref{tab:allres} for details.  While both works used
information about the near context as features, a further correction
phase was required in order to make sure that the chunk tagging yields
a proper chunking (e.g.~a `B' tag cannot follow an `O' tag). That
phase essentially makes use of information which may result from
decisions made for quite far words.

Another memory-based approach was presented by
\namecite{cardie-pierce-98}.  They created a set of grammar rules from the NP
training inventory, and pruned it using a separate corpus. In the inference
phase, the longest matching rule was applied. The system
accepts POS-tag strings, that is, it does not handle lexical
information. Direct comparison with the results of RM95 was
not presented, but cross-validation results on NPs created in a
similar fashion yielded 91.1\% recall and 90.7\% precision -- similar to
what RM95 obtained ignoring lexical data. Cardie and
Pierce have used a pruning methodology which discards the ten worst
rules until precision drops, as well as local repair heuristics which
improved the precision by 1\% without harming the recall. Their method
works better, as they note, on simpler NPs.
 
\namecite{skut-brants-98} worked also at the word-level, but with
features which include hierarchical structure information as well.
They used the maximum-entropy method for learning to assign words with
a triple structural tag: a POS tag, syntactic category, and a tag which
represents the relation between dominating nodes of the current and
the preceding words. The triple nature of the features makes it possible
to use only some of the three tags - thereby obtaining a more general
and, possibly, meaningful feature. Weights of the structural features
are evaluated using the maximum-entropy method, then incorporated into
a trigram model. The reported results for NP and PP chunking of a
German text are 88.9\% recall and 87.6\% precision.

\def\mgets{\leftarrow}

\def\OPEN{`{{\tt [\hskip-0.051cm[}}'}
\def\CLOSE{`{{\tt ]\hskip-0.051cm]}}'}

\def\Tseq#1#2{{$t_{#1},\ldots,t_{#2}$}}
\def\tup#1{{\langle{#1}\rangle}}

\def\algorithm#1#2{{\par\noindent{{\bf #1}({\it #2})}}}
\def\subhead#1{\noindent{\bf #1}\\}

\def\num{{\bf num}}
\def\minsize{{\bf minsize}}
\def\maxcontext{{\bf maxcontext}}
\def\maxleftcontext{{\bf maxleftcontext}}
\def\maxrightcontext{{\bf maxrightcontext}}
\def\maxoverlap{{\bf maxoverlap}}

\section{The Algorithm}
\label{sec:alg}

The input to the Memory-Based Sequence Learning (MBSL) algorithm is a
sentence represented as a sequence of POS tags, and its output is a
{\it bracketed sentence}, indicating which subsequences of the
sentence are to be considered instances of the target pattern ({\it
target instances}). The training corpus consists of pre-bracketed
sentences, it is used for creating the memory.
MBSL (see figure~\ref{alg:MBSL}) determines the bracketing by first
considering each subsequence of the sentence as a {\it candidate} to
be a target instance. For each candidate $c$, it computes a likelihood score
$f_C(c)$ by searching the memory, keeping $c$ if its score is above
a threshold $\th_C$.  The algorithm then finds a consistent
bracketing for the input sentence, giving preference to subsequences whose
score was high.  In the remainder of this section we will describe 
the components of the algorithm in more detail.

\vskip 1.0ex {\tt insert figure 1 here}

\begin{figure}[p]
\algorithm{MBSL}{sentence, memory, $\th_T$, $\th_C$}
\begin{enumerate}
\item Consider each subsequence of {\it sentence} as a candidate:
\item \quad Construct a situated candidate $s$ from the candidate
by prepending left context followed by \OPEN\ and appending 
\CLOSE\ followed by right context;
\item \quad Evaluate $f_C(c)=${\bf CandidateScore}($s$, {\it memory}, $\th_T$);
\item \quad If $f_C(c)>\th_C$, add $c$ to {\it candidate set};
\item {\bf SelectCandidates}({\it candidate set}).
\end{enumerate}
\caption{The top-level MBSL bracketing algorithm.\label{alg:MBSL}}
\end{figure}

\subsection{Scoring candidates}

We first describe the mechanism for scoring candidates. The input is a
candidate subsequence, along with its {\it context}, i.e.~surrounding
tags in the input sentence. We derive a score for the candidate by
searching substrings of the candidate and its context in
the training corpus.


Consider first the case when the candidate subsequence as a
whole occurs in the training corpus as an actual target instance. For
example, consider the candidate \verb+ADJ NN NN+, where the training
corpus contains the bracketed sentence
\begin{verbatim}
[ NN ] VB [ ADJ NN NN ] RB PP [ NN ] .
\end{verbatim}
Here the candidate sequence is a target instance in the
training sentence.  This fact constitutes a {\it positive} evidence that
the candidate is indeed a target instance.  The evidence would be even
stronger if the candidate's context in the input sentence was the
same as that in the training sentence.  However, if the candidate
(with or without context) also occurs in the training {\it not} as a
target instance, we would also have {\it negative} evidence 
indicating that the candidate might not be 
a target instance.  For example, the candidate \verb+NN NN RB+ 
occurs negatively in the training sentence above.

We may generalize this notion by considering positive and negative
evidence to include substrings of the candidate and its context.
For example, if the sequence \verb+ADJ NN+ often occurs at the
beginning of a noun phrase in the training, we have some
evidence that a candidate beginning with that sequence is a
noun phrase.  For example, although the candidate \verb+ADJ NN NN NN+ does not occur as a whole in the training sentence above, the sentence 
gives positive evidence for the candidate, in the form of the prefix
subsequence \verb+ADJ NN NN+ and the suffix subsequence \verb+NN NN+.

The basic idea, then, is to find the set of subsequences of the 
candidate and its context which provide positive evidence.
This set is then used to compute the candidate score.

\subsubsection{Candidates and tiles}

The MBSL scoring algorithm works by considering {\it situated
candidates}.  A situated candidate is a POS sequence
containing a pair
of brackets (\OPEN$\,\cdots\,$\CLOSE), 
indicating a candidate for a target instance.  The
portion of the sentence between the brackets is the {\it candidate} (as above),
while the portions to the left and to the right of the candidate are its
{\it context}.  

The idea of the MBSL scoring algorithm is to construct a {\it tiling} of
subsequences of a situated candidate which covers the entire
candidate.  We consider as {\it tiles} all 
subsequences of the situated candidate which contain at least a left or right
bracket.  Thus we only consider tiles
within or adjacent to the candidate that also include a candidate
boundary.  This constraint reduces the computational complexity of
the algorithm (and the run-time in practice), while 
taking into account the primary evidence for a target instance at the 
instance boundaries.

Although in principle our algorithm may work with unlimited context,
in practice we only consider a fixed amount of maximal {\it left} and
{\it right} contexts. Let the context limit be denoted by $cn$,
the total number of tiles for a candidate of length $l$ is
$n_{\rm tiles} = 2\cdot cn \cdot (l+2)+ 2l+cn^2+1$. For a fixed
maximal context, $n_{\rm tiles}=$O$(l)$, whereas for a fixed length the number
of tiles grows as $cn^2$. A large $cn$ will therefore yield an unmanageable
number of tiles. That is the reason for considering a fixed maximal context,
we have used values of 2 or 3 in our experiments.


Each tile is assigned a score based on its occurrences
in the training memory.
Since brackets correspond to boundaries of potential
target instances, it is important to consider how the bracket
positions in the tile correspond to those in the training memory.  

For example, consider again the training sentence
\begin{verbatim}
[ NN ] VB [ ADJ NN NN ] RB PP [ NN ] .
\end{verbatim}
We may now examine the occurrence in this sentence of several 
possible tiles:
\begin{description}
\item[\tt VB [ ADJ NN] occurs positively in the sentence, and
\item[\tt NN NN {]} RB] also occurs positively, while 
\item[\tt NN [ NN RB] occurs negatively in the training sentence,
since the bracket is in a different position.
\end{description}
The positive evidence for a tile is measured by its {\it positive
count}, the number of times the tile occurs in
the training memory with corresponding brackets.  Similarly, the
negative evidence for a tile is measured by its {\it negative count},
the number of times that the POS sequence of the tile occurs in the
training memory with non-corresponding brackets (either brackets in the
training are in a different position than in the tile, or without brackets).
The {\it total count} of a tile is its positive count plus its negative
count.  The score $f_T(t)$ of a tile $t$ is defined as the ratio of its
positive and total counts (other functions could also easily be considered).
\[ f_T(t) = \frac{\mbox{pos\_count}(t)}{\mbox{total\_count}(t)} \]

\comment{
The overall score of a situated candidate is a function of the scores
of all the tiles for the candidate, as well as the relations between
the tiles' positions.  These relations include tile adjacency, overlap between tiles, the amount of context in a tile, and so on.
}

We say that a tile whose score $f_T(t)$ exceeds a given threshold 
$\th_T$, and thus has `sufficient' positive evidence, is a {\it matching tile}.
Each matching tile gives supporting evidence that a part of the
candidate should be considered
part of a target instance.  In order to combine
this evidence, we try to cover the entire candidate by a set of
matching tiles, with no gaps.  Such a covering constitutes evidence
that the entire candidate is a target instance.  For example, consider
the matching tiles shown for the candidate in
figure~\ref{fig:matches}.  The set of matching tiles 2, 4, and 5
covers the candidate, as does the set of tiles 1 and 5.  Also note
that tile 1 constitutes a cover on its own.

\vskip 1.0ex {\tt insert figure 2 here}

\begin{figure}[p]
\small
\begin{verbatim}
Candidate: NN VB [ ADJ NN NN ] RB
MTile 1:      VB [ ADJ NN NN ]
MTile 2:      VB [ ADJ
MTile 3:         [ ADJ NN
MTile 4:               NN NN ]
MTile 5:                  NN ] RB
\end{verbatim}
\caption{A candidate subsequence, and 5 matching tiles found in the
training corpus.\label{fig:matches}
}
\end{figure}

To make this precise, we first say that a tile $t_1$ {\it connects} to
a tile $t_2$ if (i) $t_2$ starts after $t_1$, (ii) there is no gap
between the end of $t_1$ and the start of $t_2$ (there may be some
overlap), and (iii) $t_2$ ends after $t_1$ (neither tile includes the
other).  For example, tiles 2 and 4 in the figure connect, while tiles
2 and 5 do not, and neither do tiles 1 and 4 (since tile 1 includes
tile 4 as a subsequence).

\subsubsection{Cover statistics}

A {\it cover} for a situated candidate $c$ is a sequence of matching
tiles which collectively cover the entire candidate, including the
boundary brackets and possibly some context, such that each tile
connects to the following one.  A cover thus provides positive
evidence for the entire sequence of tags in the candidate.

The set of all covers for a candidate summarizes all of the 
evidence for the candidate as a target instance. The score of a candidate is
therefore a function of statistics of the set of all its covers.
For example, if a candidate
has many different covers, it is more likely to be a target instance, 
since many different pieces of evidence can be brought to bear.

We have empirically found several statistics of the cover set 
to be useful. These include, for each cover, the number of
matches it contains, the total number of context tags it contains, and
the number of positions which more than one match covers (the amount
of overlap).  We thus compute, for the set of all covers of a candidate $c$:
\begin{itemize}
\item The tootal number of different covers, \num($c$),
\item The least number of tiles constituting a cover, \minsize($c$),
\item The maximum amount of total context in any cover (left plus right context),
\\ \maxcontext($c$), and
\item The maximum over all covers of the total number of tile elements that
overlap between connecting tiles, \maxoverlap($c$).
\end{itemize}
Each of these items indicates the strength of the
evidence which the covers provide for the candidate.

In order to compute a candidate's statistics efficiently, we
construct the {\it cover graph} of the candidate.  The
cover graph is a directed acyclic graph (DAG) whose nodes represent
tiles of the candidate, such that an arc exists between nodes $v(t_1)$ and
$v(t_2)$ whenever tile $t_1$ connects to $t_2$.  
Two special nodes, START and END, are added to the cover graph. START is
connected to to every node whose tile contains an open bracket, and similarly,
every node representing a tile containing a close bracket is connected to END.

It is easy to see that each path from START to END constitutes a cover,
and that every cover gives such a path.
Therefore the statistics of all the covers may be efficiently
computed by a depth-first traversal of the cover graph.
An algorithm for constructing the cover graph is
given in figure~\ref{alg:covergraph}.

\vskip 1.0ex {\tt insert figure 3 here}

\begin{figure}[p]
\algorithm{CoverGraph}{situated-candidate, $\th_T$}
\begin{enumerate}
\item $V\mgets \{$START$\}$;
\item For each subsequence $t$ (potential tile) of {\it situated-candidate}
including\\ either \OPEN or \CLOSE:
\item \quad {\bf SearchTile}({\it memory, t}) to get positive and total counts;
\item \quad If $f_T(t) > \th_T$
\item \qquad add a vertex $v(t)$ to $V$;
\item $E\mgets\emptyset$;
\item For each pair of vertices $v(t_1),v(t_2)\in V$
such that $t_1$ connects to $t_2$:
\item \quad Add the arc $(v(t_1),v(t_2))$ to $E$;
\item Return the graph $(V,E)$.
\end{enumerate}
\caption{Constructing the cover graph.\label{alg:covergraph}}
\end{figure}

A graph whose structure is similar to that of the cover graph
was used by \namecite{ded-nus} and \namecite{yvon-96} for representing
alternative pronunciations of a word. In the application of \namecite{yvon-96},
the nodes are labelled with phoneme sequences and arcs connect nodes whose
labels overlap. START and END nodes are connected to prefix and suffix nodes
respectively; each possible pronunciation thus corresponds to a path from
START to END. Nevertheless, the scoring mechanism is different as the aim
is to find a best path rather than calculating statistics of a graph.
Similar with candidate score, though, the path scoring function of
\namecite{yvon-96} also prefers more overlap and short paths.

\subsubsection{Computing the candidate score}

The score of the candidate is a linear function of its cover statistics:
\[
\begin{array}{rcl}
   f_C(c) &=& \alpha\,\num(c) - \beta\,\minsize(c) +\\
	&&  \gamma\,\maxcontext(c) + \\
	&&	\delta\,\maxoverlap(c)
\end{array}
\]
If candidate $c$ has no covers, $f_C(c)=0$.  Note that
\minsize\ is weighted negatively, since a cover with fewer tiles
provides stronger evidence for the candidate. The candidate scoring
algorithm is given in figure~\ref{alg:candscore}.

In the current implementation, the weights were chosen so as to give a
lexicographic ordering on the features, 
preferring first candidates with more covers,
then those with covers containing fewer tiles, 
then those with larger covered contexts, and
finally, when all else is equal, preferring candidates whose covers
have more overlap between connecting tiles. 
We plan to investigate in the future a data-driven approach
(based on the Winnow algorithm \cite{Litt88}) for optimal selection and
weighting of statistical features of the score.

\vskip 1.0ex {\tt insert figure 4 here}

\begin{figure}[p]
\algorithm{CandidateScore}{situated-candidate,memory, $\th_T$}
\begin{enumerate}
\item Let $G$ = {\bf CoverGraph}({\it situated-candidate}, $\th_T$);
\item Compute \num, \minsize, \maxcontext, and \maxoverlap\ by performing DFS on $G$;
\item Return the candidate score $f_C$ as a function of these statistics.
\end{enumerate}
\caption{The candidate scoring algorithm\label{alg:candscore}}
\end{figure}

\subsubsection{Complexity}

We analyse the worst-case complexity of creating the cover graph and 
computing the score of a situated candidate in terms of the candidate length
$l$. The steps of that process are:
\begin{description}
\item[Create vertices:] Since there are O($l$) potential
tiles $t$ in the situated candidate, and searching for a tile in the memory takes 
linear time (see below in section~\ref{sec:memory}), 
this step could take O($l^2$).  
\item[Create edges:] There are at most O($l^2$) possible edges in the graph,
so this step takes O($l^2$).
\item[Computing statistics:] The DFS takes O($|V|+|E|$)=O($l^2$).
\item[Computing the score:] This takes a constant time.
\end{description}
Hence computing the candidate score for a situated candidate takes 
O($l^2$) in the worst case.  In practice, however, this worst case is rarely
reached, since usually only a small fraction of the tiles are matching.

\subsection{Selecting candidates}

In order to select a bracketing for the input sentence, we
assume that target instances are non-overlapping (this is the case for
the types of linguistic patterns with which we experimented). To apply this
constraint we use a simple constraint propagation algorithm that finds
the best choice of non-overlapping candidates in an input sentence, given
in figure~\ref{alg:selection}.
Other methods for determining the preferred bracketing may also be 
applied; this remains a topic for future research.

\vskip 1.0ex {\tt insert figure 5 here}

\begin{figure}[p]
\algorithm{SelectCandidates}{candidate-set}
\begin{enumerate}
\item Examine each candidate $c\in${\it candidate-set} such that $f_C(c) > 0$, in descending order of $f_C(c)$: 
\item \quad Add $c$'s brackets to the sentence;
\item \quad Remove all candidates overlapping with $c$ from {\it candidate set};
\item Return the candidates remaining in {\it candidate-set} as target instances.
\end{enumerate}
\caption{Candidate selection algorithm.\label{alg:selection}}
\end{figure}

We analyse the complexity of candidate selection by first noting
that the number of candidates in the candidate set is at most O($n^2$) where $n$ is the length of the input sentence.  
The first step in selection is to sort the candidate set by
$f_C$, which takes O($n^2\log{n^2}$)= O($n^2\log{n}$).
Since there are at most
O($n$) candidates in the final bracketing, with proper indexing 
removing overlapping candidates takes a total of at most O($n^2$) time, giving a total time of O($n^2\log{n}$).

\subsection{Worst-case complexity of MBSL}

The overall complexity of the bracketing algorithm
(figure~\ref{alg:MBSL}) may now be easily computed.  For a sentence of
length $n$, there are O($n^2$) candidates, each of length at most
O($n$).  Hence, since scoring a single candidate of length $l$ takes O($l^2$), 
scoring all of the candidates takes at most O($n^4$).
Since the selection step takes O($n^2\log{n}$), the worst case
complexity of the MBSL bracketing algorithm is O($n^4$).
In practice, however, this worst-case is rarely reached.

\subsection{Implementing the training memory}
\label{sec:memory}

The MBSL scoring algorithm above needs to search the training corpus for many
subsequences of each input sentence in order to find matching tiles.
Implementing this search efficiently is therefore of prime importance.
We do so by encoding all possible tiles for each target instance (positive
examples) in the training corpus using a trie data structure.
A trie is a tree whose arcs are labeled with 
characters such that each path from the root represents a distinct tile
in the training.  A trie allows searching for a sequence
in time linear in the length of the sequence.
Each node in the trie represents the sequence of 
arc labels along the path reaching it from the root.
We store the positive and total counts  for each tile at its associated node
in the trie.  

Given a trie as described, 
determining the positive and total counts of a given
potential tile is done by searching the trie for the tile, and simply returning
the counts stored at the node found.  This takes time linear in the length
of the potential tile.  If tile $t_1$ is a prefix of $t_2$, then searching for 
$t_2$ after searching for $t_1$ may be done incrementally, by starting from $t_1$'s node.  Hence searching for a set of tiles all starting at the same 
point in the text is performed incrementally in linear (rather than quadratic) time.

The memory trie itself is built in two passes. Their worst-case
complexity depends on $k$, the number of sentences in the corpus, 
and $n$, the maximum sentence length.

In the first pass, all the tiles of each target instance
in the corpus are inserted into the trie.
For example, considering a maximum context of 1, given the NP instance
\begin{verbatim}
 VB [ NN ] IN
\end{verbatim}
the possible tiles are:
\begin{verbatim}
 VB [
 VB [ NN
 VB [ NN ]
 VB [ NN ] IN
 [ NN
 [ NN ]
 [ NN ] IN
 NN ]
 NN ] IN
  ] IN
\end{verbatim}
For each possible tile $t$, if $t$ is not in the trie, new nodes required
for constructing a path representing $t$ are added, and
the positive count of the final node of the path is set to 1. Otherwise, if
$t$ is already in the trie, the positive count of its associated node is
increased by 1.
This search is performed incrementally for tiles sharing a prefix and so 
adding all tiles starting at a given point in a target instance takes 
time linear in the length of the tile (at most O($n$)).
Thus, since there are at most O($n$) tile starting points in a sentence, 
the first pass takes time at most O($kn^2$).

In the second pass, we compute the total counts for each node by
considering all the subsequences of every sentence in the corpus.  For
each such subsequence $s$, we find each node in the trie whose
associated subsequence is identical to $s$ except possibly for the
addition of brackets.  We then increment the total count for
each such matching node.  Since each tile is a substring of a sentence,
there are at most O($n^2$) ways to construct a possible tile for $s$.
Therefore, there are at most O($n^2$) matching nodes in the trie.  Here too we
search the trie for all subsequences that share a starting point
incrementally, and so the search takes time O($n^3$) for each starting
point, and O($kn^4$) for the entire corpus.  
Therefore the overall worst-case
complexity of building the memory is O($kn^4$).  Although this
complexity looks large, in practice the worst case is not really
reached. The reason is that pattern instances, hence tiles, are commonly much
shorter than the sentence in which they are embedded. The actual time
required for building the memory is not long, for example:
in the experiments on NPs, the training data contain 8936 sentences, 54760
patterns, and 229,598 words. With a maximum context of 3, it took 16 seconds
to build the memory on a 300 MHz Pentium II running Linux.
The bracketing rate was 1000 sentences per minute, and the trie took
up 100MB of memory.

In a previous implementation \cite{ADK98} we encoded the entire
corpus in two suffix trees. Inspired by \namecite{sh-97}, one suffix tree
encoded the positive examples and the other encoded the entire corpus, and was
used for obtaining the total counts of tiles.
Following \namecite{roth-pc} we have decided to use a trie of positive tiles
only. While using suffix trees gives a better worst-case time complexity for
building the memory, using a trie improves performance by 25-30\%.  This is
mainly because the trie is used for storing positive examples only.

\section{Evaluation}
\label{sec:eval}





\subsection{The Data}

We have tested our algorithm in recognizing three syntactic patterns:
noun phrase sequences (NP), verb-object (VO), and subject-verb (SV) relations.
The training and testing data were derived from the Penn TreeBank.

We used the NP data prepared by RM95, these data were tagged by Brill's
POS tagger. 
The SV and VO data were extracted using T (TreeBank's search script
language) scripts (available at {\tt
http://www.cs.biu.ac.il/$\sim$yuvalk/MBSL}). Table~\ref{tab:size}
summarizes the sizes of the training and test data sets and the number
of examples in each.  The T scripts did not attempt to match
dependencies over very complex structures, since we are concerned with
shallow, or local, patterns.

\vskip 1.0ex {\tt insert table 1 here}

\begin{table}[p]
\center{Train Data:}\\
\begin{tabular}{|c|c|c|c|}
 & sentences & words & patterns \\
\hline
 NP &  8936  & 229598 & 54760 \\
 VO & 16397  & 454375 & 14271 \\
 SV & 16397  & 454375 & 25024 \\
\hline
\end{tabular}
\center{Test Data:}\\
\begin{tabular}{|c|c|c|c|}
 & sentences & words & patterns \\
\hline
 NP & 2012 & 51401 & 12335 \\
 VO & 1921 & 53604 & 1626 \\
 SV & 1921 & 53604 & 3044 \\
\hline
\end{tabular}
\caption{Sizes of training and test data\label{tab:size}
}
\end{table}

For VO patterns, we put the starting delimiter just before the main
verb and the ending delimiter just after the object head.
For example:
\begin{verbatim}
 ... investigators started to
 [ view the lower price levels ]
 as attractive ...
\end{verbatim}
We used a similar policy for SV patterns, defining the start of the pattern at
the start of the subject noun phrase and the end right at the first verb
encountered.  For example:
\begin{verbatim}
 ... argue that
 [ the U.S. should regulate ]
 the class ...
\end{verbatim}
The subject and object noun-phrase boundaries were those specified by the
Penn TreeBank annotators, and phrases containing conjunctions or appositives
were not further analysed.

By terminating the SV pattern right after the first verb encountered
we get seemingly incorrect relation descriptions such as:
\begin{verbatim}
 [ the bird which stands ] there sings
\end{verbatim}
While the actual SV pair is {\tt bird - sings}, the T script extracts
{\tt bird - stands}. This pair of words is not a properly grammatical SV,
but it contains important information and may be regarded as
grammatically meaningful in its own right. Hence we do not believe
that it impinges on the evaluation of our learning method.

Table~\ref{tab:lhist} shows the distribution of pattern length in the train
data. We did not attempt to extract passive-voice VO relations.  We
have extracted SVs from 92.4\% of the sentences and VOs from
58.5\% (these figures refer to the overall train and test
data). Almost all (99.8\%) of the sentences in the NP data contain a
noun-phrase. Table~\ref{tab:nwne} shows the number of examples for corpus
sizes of 50K, 100K, 150K, and 200K.

\vskip 1.0ex {\tt insert table 2 here}

\begin{table}[p]
\begin{center}
\begin{tabular}{|c|r|r|r|r|r|r|}
 Len & NP & \% & VO & \% & SV & \% \\
 \hline
  1 & 16959 & 31 &       &    &       &    \\
  2 & 21577 & 39 &  3203 & 22 &  7613 & 30 \\
  3 & 10264 & 19 &  5922 & 41 &  7265 & 29 \\
  4 &  3630 &  7 &  2952 & 21 &  3284 & 13 \\
  5 &  1460 &  3 &  1242 &  9 &  1697 &  7 \\
  6 &   521 &  1 &   506 &  4 &  1112 &  4 \\
  7 &   199 &  0 &   242 &  2 &   806 &  3 \\
  8 &    69 &  0 &   119 &  1 &   592 &  2 \\
  9 &    40 &  0 &    44 &  0 &   446 &  2 \\
 10 &    18 &  0 &    20 &  0 &   392 &  2 \\
 $>$10 & 23 &  0 &    23 &  0 &  1917 &  8 \\
\hline
total  & 54760 & & 14271 &    & 25024 &    \\
\hline
avg. len& 2.2 &  &  3.4   &    & 4.5   &    \\
\hline
\end{tabular}
\caption{Distribution of pattern lengths, total number of patterns and
         average length in the training data.\label{tab:lhist}}
\end{center}
\end{table}

\vskip 1.0ex {\tt insert table 3 here}

\begin{table}[p]
\centering
\begin{tabular}{|c|c|c|c|}
 Words & NP & VO & SV \\
 \hline
  50K & 12100 & 1603 & 2764 \\
 100K & 23864 & 3180 & 5385 \\
 150K & 35799 & 4798 & 8237 \\
 200K & 47730 & 6313 & 11034 \\
\hline
\end{tabular}
\caption{Number of pattern instances within 50K to 200K words\label{tab:nwne}
}
\end{table}




\subsection{Testing Methodology}


The test procedure has two parameters: (a) maximum context size of a
candidate, which limits {\em what} queries are performed on the
memory, and (b) the threshold $\th_T$ used for establishing a matching
tile, which determines {\em how} to make use of the query results.
(The candidate score threshold, $\th_C$ was set to 0, such that every candidate
with at least one cover was considered.)


Recall and precision figures were obtained for various parameter values.
$F_\beta$, a common measure in information retrieval \cite{vanrijs-79},
was used as a single-figure measure of performance:
\[
  F_\beta ={{(\beta^2+1)\cdot P \cdot R} \over {\beta^2\cdot P + R}}
\]
We use $\beta$ which gives no preference to either
recall or precision.

Performance may be measured also on a word-by word basis, counting as a
success any word which was identified correctly as being part of the target
pattern. That method was employed, along with recall/precision, by RM95.
We preferred to measure performance by recall and precision for complete
patterns. Most errors involved identifications of slightly shifted, shorter or
longer sequences. Given a pattern consisting of five words, for
example, identifying only a four-word portion of this pattern would
yield both a recall and precision errors. Tag-assignment scoring, on the
other hand, will give it a score of 80\%. We hold the view that such
an identification is an error, rather than a partial success.



\subsection{Results}
\label{sec:res}

Table~\ref{tab:allres} summarizes the optimal parameter settings and results for NP,
VO, and SV on the test set. In order to find the optimal values of the
context size and threshold, we tried $ 0.1 \le \th_T \le 0.95$, and
maximum context sizes of 1,2, and 3. Our experiments used 5-fold
cross-validation on the training data to determine the optimal
parameter settings.

\vskip 1.0ex {\tt insert table 4 here}

\begin{table}[p]
\centering
\begin{tabular}{|c|c|c|c|c|c|c|ll}
  & Con. & Thresh. & Break  Even & Recall (\%) & Precision (\%)& $F_{\beta=1}$\\
\hline
 VO &  2 & 0.5 & 81.3 & 89.8  & 77.1 & 83.0 \\
 SV &  3 & 0.6 & 86.1 & 84.5  & 88.6 & 86.5 \\
 NP &  3 & 0.6 & 91.4 & 91.6  & 91.6 & 91.6 \\
 NP$^\dagger$ &  3 & 0.6 &  & 92.4  & 92.4 & 92.4 \\ 
\hline
\hline
RM95 (NP) & - & - & -   & 90.7 & 90.5 & 90.6 \\
   + Lex & - & - & -   & 92.3 & 91.8 & 92.0 \\
Veenstra (NP) & - & - & -   & 94.3 & 89.0 & 91.6 \\
\hline
\end{tabular}
\caption{Results with optimal parameter settings for context size
and threshold, and breakeven points. The NP$^\dagger$ line presents results
obtained when removing adverbs and prepositions from the beginning of NPs
(section~\ref{sec:res}). The last two lines show the
results of RM95 (with and without lexical features) and Veenstra
(1998, using a different POS tagger)
recognizing NPs results with the same train/test data.
The optimal parameters were obtained by 5-fold cross-validation. Note that the
results for SV and VO were obtained from Penn TreeBank data which include
trace markers. In table~\ref{tab:res-more} we present results without
the trace markers. These results are degraded, especially for VO.
\label{tab:allres}
}
\end{table}

In experimenting with the maximum context size parameter, we found that
the difference between the values of $F_{\beta}$ for context sizes of 2 and
3 is less than 0.5\% for the optimal
threshold. Scores for a context size of 1 yielded $F_{\beta}$ values
smaller by more than 1\% than the values for the larger contexts.





\begin{figure}[p]
\begin{center}
\includegraphics[width=3.13in, height=3.13in]{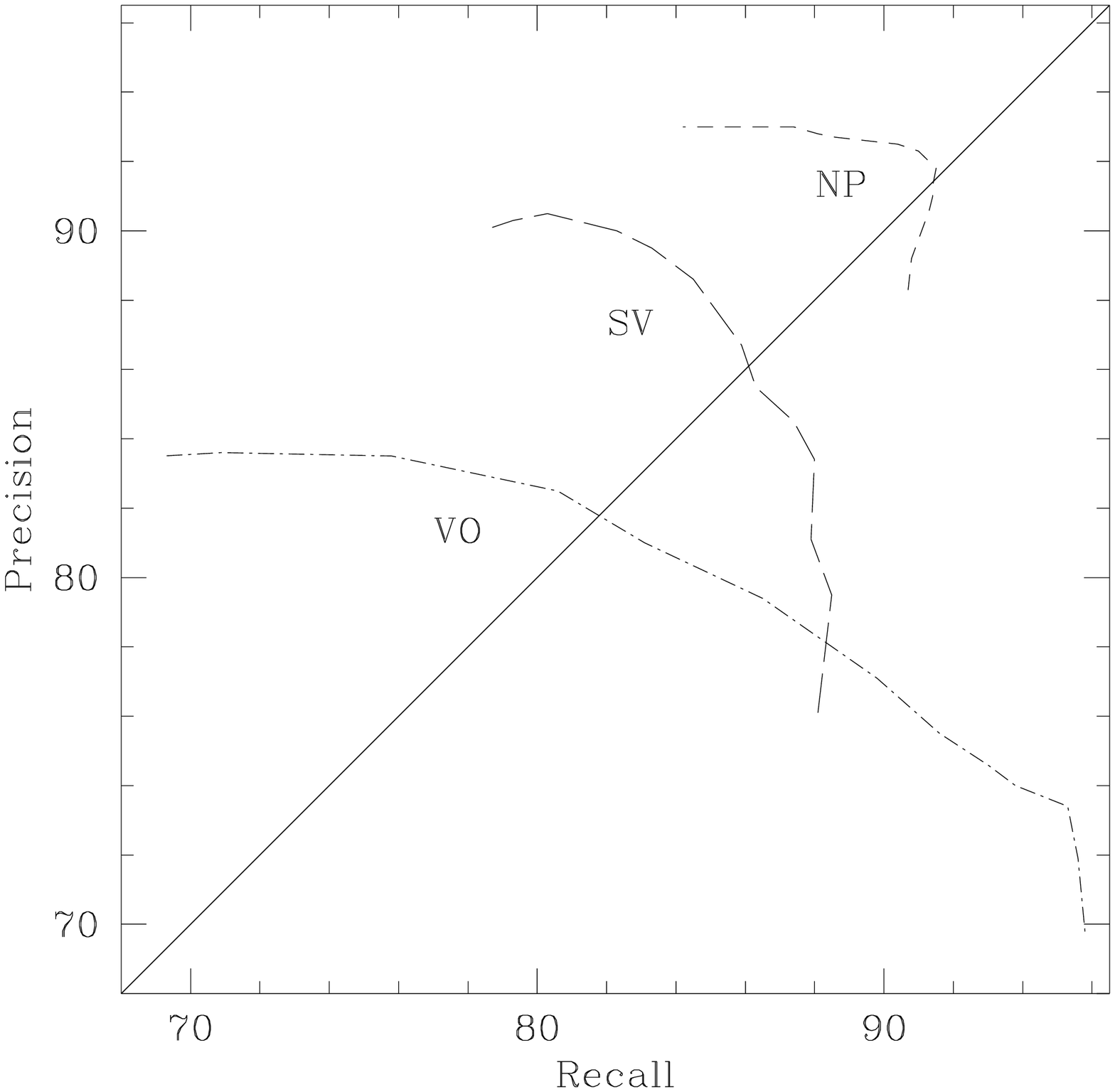}
\end{center}
\caption{Recall-Precision curves for NP, VO, and SV; $0.1 \le \th \le 0.99$
  \label{fig:recprec}
}
\end{figure}

Figure~\ref{fig:recprec} shows recall/precision curves for the three data sets,
obtained by varying $\th_T$ while keeping the maximum context size at
its optimal value. The difference between $F_{\beta=1}$ values for
different thresholds was always less than 2\%.

\vskip 1.0ex {\tt insert figure 6 here}

\begin{figure}[p]
\includegraphics[height=2.65in,width=3.0in]{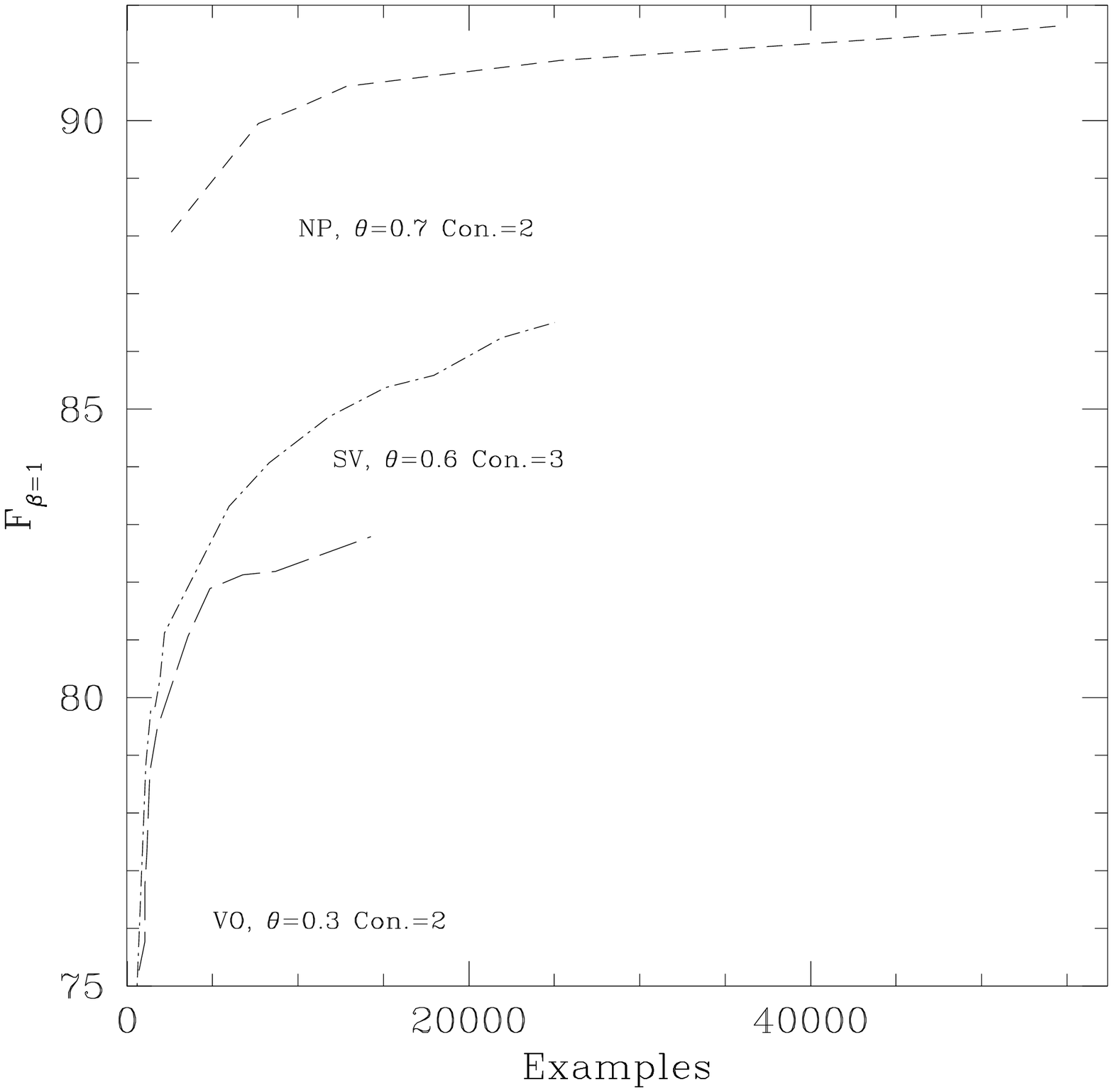}
\hskip 0.4in
\includegraphics[height=2.65in,width=3.0in]{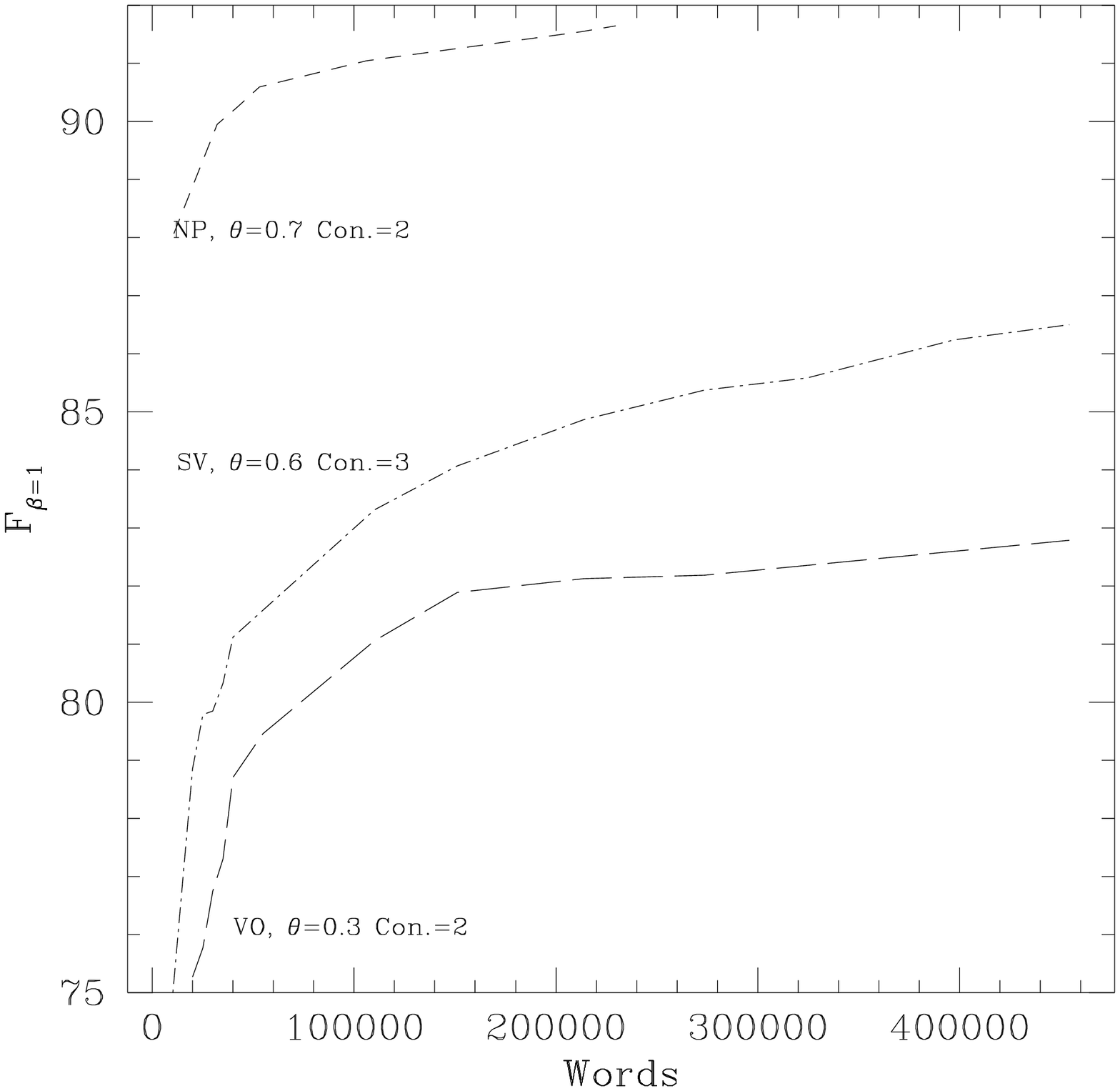}
\caption{Learning curves for NP, VO, and SV by number of examples (left)
and words (right)\label{fig:lc}}
\end{figure}

Figure~\ref{fig:lc} shows the learning curves by
amount of training examples and number of words in the training data,
for particular parameter settings.

\vskip 1.0ex {\tt insert figure 7 here}

The results of RM95 are shown in table~\ref{tab:allres} for comparison,
along with those of \namecite{veenstra-np-98}.
All the cited results pertain to a training set of 229,000 words.
Using a larger training set, of 950,000 words, RM95 attained a recall/precision
of 93.5\%/93.1\% ($F_\beta$=93.3\%); this last result was achieved including lexical information.

Two results from RM95 are presented: for learning with and without
lexical information.  Only the latter should be compared with the
results of our algorithm, because it also does not consider lexical
information.  All results for NPs are quite similar, up to recall/precision
tradeoff. That may be related to limitations inherent in the
particular choice of train/test data.

Some of the errors resulted from
noun-phrases beginning with a preposition, or an adverb, such as:
\begin{verbatim}
[ IN About CD 20,000 NNS years ] RB ago

[ RB yet DT another NN setback ]
\end{verbatim}
When we focus on nouns, preceding
prepositions and adverbs are not of interest. We have made an experiment
in which these words were taken out of NPs (and NPs composed of one word
tagged {\tt IN} or {\tt RB} (e.g.~{\tt IN that}) were ignored). This
experiment yielded better results, as  presented
in table~\ref{tab:allres} at the line marked NP$^\dagger$.

The Penn TreeBank data contain trace markers, tagged as {\tt -NONE-}. These would
not appear in a raw text from another source. In order to see the effect of the
trace markers we have removed them and experimented with the optimal
parameters. The results are summarized in table~\ref{tab:res-more}. The more
poor results indicate that the trace markers actually improved the inference
quality, especially for the VO pattern.

Both RM95 and Veenstra approach the chunking task as a classification
task, using chunk-tags. Both works use chunk-tag assignments of nearby (up to
a distance of 2) words as features, and employ a post-processing step for
assuring that the chunk-tags yield proper chunking (for example, correct cases
where a chunk is opened within another chunk). Since chunk tags of preceding
words were obtained using information about even earlier words - using
them as features implies that information about preceding words at larger
distances is taken into account. In these approaches there is an asymmetry
between words to the left, and words to the right of the current word,
as tagging is carried out from left to right.

Our method differs from these methods in that it considers sequences
rather than individual words. Tiles of any length may contribute to the
decision, with no preference to the beginning or ending of a pattern
(in contrast with \namecite{cardie-pierce-98} who used only the prefixes
of NPs). That makes the presented algorithm distinct from finite-state
methods, which are directional in nature.

Indeed, we tried to learn a stochastic transducer at an earlier phase of the
research, using the method of \namecite{rst-apfa-95}. We trained the transducer
for detecting NPs in POS sequences, with the sentence presented at
different offsets so the transducer could identify different NPs.
That method, however, yielded poor results, possibly due to the
difficulty of combining generalizations about both the beginning of an NP
and its end.

\subsection{Common Errors}
\label{sec:err}

This section contains a brief account of typical error-sources for the
 NP, SV, and VO learning tasks. In the examples shown {\tt [ ]} bracket
 the true pattern instance, whereas {\tt [[ ]]} bracket the one found
 by the algorithm.

The NP data were tagged automatically, that certainly introduced some
errors. Since the tagging errors are consistent throughout the train and
test data, their effect is to increase data sparseness.
Most of the errors in the NP learning task were due to:
\begin{itemize}

\item Coordinations and punctuation marks: Depending on the context, some NPs
 which contain a coordination were split and some NPs separated by a
 coordination were detected as a single NP.
\begin{verbatim}
, , [ [[ DT a JJ local NN lawyer ]] CC and
[[ JJ human-rights NN monitor ]] ], ,


NN relationship IN between [[ [ NNP Mr. NNP Noriega ]
CC and NNP [ Washington ] ]] IN that
\end{verbatim}

In order to estimate the effect of coordinations we made a test in which
all NPs containing {\tt CC}'s were split to their more basic components.
That increased the number of phrases by 2\%, and the recall/precision to
93.4\% and 92.9\% respectively. However, we are then faced with the new
problem of deciding whether or not to create a single NP from two NPs
connected by a coordination.
\item Adjacent phrases: Cases where two adjacent NPs were detected as a
single phrase
\begin{verbatim}
VBZ recalls [[ [ PRP$ his NN close NN ally ]
[ NNP Jose NNP Blandon ] ]] , ,
\end{verbatim}
\item Ambiguity of verbs: Words tagged {\tt VBG} or {\tt VBN} may be verbs
as well as modifiers. The way these words are handled depends on the context
\begin{verbatim}
  , , [VBN fixed [[ VBG leading NNS edges]] ] IN for 
\end{verbatim}
Note that {\tt leading} was part of the detected NP, probably due to the
{\tt VBN} context, vs.
\begin{verbatim}
PRP it VBD had [VBG operating [[ NN profit ]] ] IN of $ $ CD million
\end{verbatim}
where the context is different.
A proper treatment of this issue has to take into account lexical information,
not only POS data.
\item Quotation marks: Many of the errors involved phrases which contain
quotation marks
\begin{verbatim}
IN by [ `` `` [[ DT a RB very JJ meaningful '' ''
NN increase ]] ] IN in
\end{verbatim}
A test with all quotation marks completely removed have not yielded
any significant improvement in the results.
\end{itemize}

In the SV and VO learning tasks, the text was taken directly from Penn
TreeBank, hence the number of tagging errors was smaller. We have
made tests with quotation marks removed for these tasks as well, using
the optimal parameters.
The result was a degradation of the precision by 1\% for SV and 0.5\% for VO,
along with a slight increase of the recall (0.2\% for SV and 0.4\% for VO).
These effects are insignificant.

The VO patterns did not take in account cases where the verb is a be-verb,
such as the pair {\tt is-chairman} in
\begin{verbatim}
NNP Mr. NNP Vinken VBZ is NN chairman IN of
\end{verbatim}
Using POS tags only, it is impossible to distinguish these verbs. We therefore
conducted an experiment in which {\tt be, am, is, are, were,} and {\tt was}
(and their abbreviations) were tagged with the new POS tag {\tt VBE}.
That was a rudimentary way of introducing lexical knowledge. The results
of this experiment are
presented in table~\ref{tab:res-more}. The incorporated knowledge caused an
improvement of 1.5\% in the recall and a more significant precision improvement
of 4.3\%$\,$.

A significant part of the remaining VO recognition errors were results of verbs
appearing as modifiers rather than actions:
\begin{verbatim}
NN asbestos [[ VBG including NN crocidolite ]] RBR more RB stringently

DT all [[ VBG remaining NNS uses ]] IN of

TO to [ VB indicate [[ VBG declining NN interest NNS rates ]] ]
\end{verbatim}

Regarding SV learning, most of the recall errors involve very long
pattern instances. As table~\ref{tab:lhist} shows, about 8\% of the SV patterns
are more than 10 words long. With such length, the diversity is large and
it is harder to model all the cases. Similarly, many of the precision
mistakes are a result of erroneously detecting relatively short instances,
or sub-parts of a pattern. The variety of possible tiles increases the chance
of false detections.

\vskip 1.0ex {\tt insert table 5 here}

\begin{table}[p]
\centering
\begin{tabular}{|c|c|c|c|c|c|ll}
  & Con. & Thresh. & Recall (\%) & Precision (\%)& $F_{\beta=1}$\\
\hline
 VO &  2 & 0.5 & 87.6 & 69.2 & 77.3 \\
 VO$^{\rm Be}$ &  2 & 0.5 & 89.2 & 73.6 & 80.6 \\
 SV &  3 & 0.6 & 84.0 & 87.5 & 85.7 \\
\hline
\end{tabular}
\caption{SV and VO results without trace markers, notice the degradation
compared with the original data. The line labelled VO$^{\rm Be}$ presents
results obtained with all be-verbs tagged by a special POS tag.
\label{tab:res-more}
}
\end{table}

\section{Conclusions}
We have presented a novel general algorithm  for learning sequential patterns.
Applying the method to three syntactic patterns in English yielded positive
results, suggesting its applicability for recognizing syntactic chunks.
The results for noun-phrase learning are compatible with current state
of the art.
The algorithm achieved good results also when applied to patterns which
constitute a relation between words (subject-verb, verb-object),
rather than a chunk.

Many of the errors can be attributed to the lack of lexical and semantic
information, as the input consists only of POS tags. These errors point out
sub-problems (e.g.~ambiguity of gerunds) which cannot be handled using
POS information alone.

The presented algorithm is part of a more comprehensive learnable shallow
parser under plan. The planned shallow parser will handle sequential patterns
(chunks) as well
as attachment problems (e.g.~prepositional phrase attachment). Attachment
problems, like many of the cases where POS data are not enough, require
lexical information. We plan to use
SNOW \cite{roth-snow}, a feature-based network learning architecture based on
Winnow, for learning to handle such problems.
SNOW can also be used within our algorithm, for a principled candidate scoring
through weighting the statistical features of candidates.

\section{acknowledgements}
The authors wish to thank Yoram Singer for his
collaboration in an earlier phase of this research project, and Giorgio Satta
for helpful discussions. We also thank Jorn Veenstra, Min Yen Kan,
and the anonymous reviewers for their instructive comments.
This research was supported in part by grant 498/95-1 from the
Israel Science Foundation, and by grant 8560296 from the Israeli Ministry
of Science.

\bibliographystyle{jetai}
\bibliography{paper}

\begin{thebibliography}{}

\bibitem[\protect\citename{Abney}1991]{Abney91}
Abney, S.~P.,
\newblock 1991,
\newblock Parsing by chunks,
\newblock In R.~C. Berwick, S.~P. Abney, and C.~Tenny, editors, {\em
  Principle-Based Parsing: Computation and Psycholinguistics}, Kluwer,
  Dordrecht, pp. 257--278.

\bibitem[\protect\citename{A{\"\i}t-Mokhtar and
  Chanod}1997]{ait-mokhtar-chanod-97}
A{\"\i}t-Mokhtar, S., and Chanod, J.-P.,
\newblock 1997,
\newblock Subject and object dependency extraction using finite-state
  transducers,
\newblock In {\em ACL'97 Workshop on Information Extraction and the Building of
  Lexical Semantic Resources for NLP Applications}, Madrid.

\bibitem[\protect\citename{Appelt \bgroup et al.\egroup }1993]{Appelt:93IJ}
Appelt, D.~E., Hobbs, J.~R., Bear, J., Israel, D., and Tyson, M.,
\newblock 1993,
\newblock Fastus: A finite-state processor for information extraction from
  real-world text,
\newblock In {\em Proc. of the 13th IJCAI}, pp. 1172--1178, Chambery, France.

\bibitem[\protect\citename{Argamon, Dagan, and Krymolowski}1998]{ADK98}
Argamon, S., Dagan, I., and Krymolowski, Y.,
\newblock 1998,
\newblock A memory-based approach to learning shallow natural language
  patterns,
\newblock In {\em Proc.~of COLING/ACL}, pp. 67--73, Montreal, Canada.

\bibitem[\protect\citename{Bod}1992]{Bod92CO}
Bod, R.,
\newblock 1992,
\newblock A computational model of language performance: Data oriented parsing,
\newblock In {\em Coling}, pp. 855--859, Nantes, France.

\bibitem[\protect\citename{Brill}1992]{Brill92}
Brill, E.,
\newblock 1992,
\newblock A simple rule-based part of speech tagger,
\newblock In {\em proc.~of the DARPA Workshop on Speech and Natural Language}.

\bibitem[\protect\citename{Brill}1993]{Brill93}
Brill, E.,
\newblock 1993,
\newblock Automatic grammar induction and parsing free text: A
  transformation-based approach,
\newblock In {\em Proc.~of the Annual Meeting of the ACL}.

\bibitem[\protect\citename{Briscoe and Carroll}1993]{bris-car-93}
Briscoe, T., and Carroll, J.,
\newblock 1993,
\newblock Generalized probabilistic {LR} parsing of natural language corpora
  with unification-based grammars,
\newblock {\em Computational Linguistics}, 19(1):25--60.

\bibitem[\protect\citename{Cardie}1993]{cardie93}
Cardie, C.,
\newblock 1993,
\newblock A case-based approach to knowledge acquisition for domain-specific
  sentence analysis,
\newblock In {\em Proceedings of the 11th National Conference on Artificial
  Intelligence}, pp. 798--803, Menlo Park, CA, USA, AAAI Press.

\bibitem[\protect\citename{Cardie and Pierce}1998]{cardie-pierce-98}
Cardie, C., and Pierce, D.,
\newblock 1998,
\newblock Error-driven pruning of treebank grammars for base noun phrase
  identification,
\newblock In {\em Proc.~of COLING/ACL}, pp. 218--224, Montreal, Canada.

\bibitem[\protect\citename{Chitrao and Grishman}1990]{chit-grish-90}
Chitrao, M., and Grishman, R.,
\newblock 1990,
\newblock Statistical parsing of messages,
\newblock In {\em Proceedings of DARPA Speech and Natural Language Processing},
  Morgan Kaufman: New York.

\bibitem[\protect\citename{Church}1988]{Church88}
Church, K.~W.,
\newblock 1988,
\newblock A stochastic parts program and noun phrase parser for unrestricted
  text,
\newblock In {\em proc.~of ACL Conference on Applied Natural Language
  Processing}.

\bibitem[\protect\citename{Collins}1997]{collins-97}
Collins, M.,
\newblock 1997,
\newblock Three generative, lexicalised models for statistical parsing,
\newblock In {\em Proc.~of the ACL/EACL Annual Meeting}, Madrid, Spain.

\bibitem[\protect\citename{Daelemans \bgroup et al.\egroup
  }1996]{daelemans-mb-96}
Daelemans, W., Zavrel, J., Berck, P., and Gillis, S.,
\newblock 1996,
\newblock {MBT}: A memory-based part of speech tagger generator,
\newblock In Eva Ejerhed and Ido Dagan, editors, {\em Proceedings of the Fourth
  Workshop on Very Large Corpora}, pp. 14--27, ACL SIGDAT.

\bibitem[\protect\citename{Daelemans, van~den Bosch, and Weijters}1996]{igtree}
Daelemans, W., van~den Bosch, A., and Weijters, T.,
\newblock 1996,
\newblock {IGTree}: Using trees for compression and classification in lazy
  learning algorithms,
\newblock {\em D. Aha ed., Artificial Intelligence Review, special issue on
  Lazy Learning}.

\bibitem[\protect\citename{{Dedina} and {Nusbaum}}1991]{ded-nus}
{Dedina}, M.~J., and {Nusbaum}, H.~C.,
\newblock 1991,
\newblock {PRONOUNCE}: a program for pronunciation by analogy,
\newblock {\em Computer Speech and Langage}, 5:55--64.

\bibitem[\protect\citename{Eisner}1996]{eisner-96}
Eisner, J.,
\newblock 1996,
\newblock Three new probabilistic models for dependency parsing: An
  exploration,
\newblock In {\em Proc.~of COLING}, pp. 340--346, Copenhagen, Denmark.

\bibitem[\protect\citename{Gee and Grosjean}1983]{gee-grosjean}
Gee, J.~P., and Grosjean, F.,
\newblock 1983,
\newblock Performance structures: A psycholinguistic and linguistic appraisal,
\newblock {\em Cognitive Psychology}, 15:411--458.

\bibitem[\protect\citename{Greffenstette}1993]{gref:acl93}
Greffenstette, G.,
\newblock 1993,
\newblock Evaluation techniques for automatic semantic extraction: Comparing
  syntactic and window based approaches,
\newblock In {\em {ACL} Workshop on Acquisition of Lexical Knowledge From
  Text}, Ohio State University.

\bibitem[\protect\citename{Littlestone}1988]{Litt88}
Littlestone, N.,
\newblock 1988,
\newblock Learning quickly when irrelevant attributes abound: A new
  linear-threshold algorithm,
\newblock {\em Machine Learning}, 2:285--318.

\bibitem[\protect\citename{Magerman and Marcus}1991]{mag-mar-91}
Magerman, D., and Marcus, M.,
\newblock 1991,
\newblock Pearl: {A} probabilistic chart parser,
\newblock In {\em Fourth DARPA Workshop on Speech and Natural Language},
  Pacific Grove, CA.

\bibitem[\protect\citename{Magerman}1995]{magerman95}
Magerman, D.~M.,
\newblock 1995,
\newblock Statistical decision-tree models for parsing,
\newblock In {\em Proc.\ of the 33{\/$^{rd}$} Annual Meeting of the Association
  for Computational Linguistics. Cambridge, {MA}, 26--30}.

\bibitem[\protect\citename{Marcus, Santorini, and
  Marcinkiewicz}1993]{TB_corpus}
Marcus, M.~P., Santorini, B., and Marcinkiewicz, M.,
\newblock 1993,
\newblock Building a large annotated corpus of {E}nglish: The {P}enn
  {T}reebank,
\newblock {\em Computational Linguistics}, 19(2):313--330.

\bibitem[\protect\citename{Pereira and Schabes}1992]{per-schabes-92}
Pereira, F., and Schabes, Y.,
\newblock 1992,
\newblock Inside-outside reestimation from partially bracketed corpora,
\newblock In {\em Proc.~of the Annual Meeting of the ACL}.

\bibitem[\protect\citename{Ramshaw and Marcus}1995]{ram-mar-95}
Ramshaw, L.~A., and Marcus, M.~P.,
\newblock 1995,
\newblock Text chunking using transformation-based learning,
\newblock In {\em Proceedings of the Third Workshop on Very Large Corpora}.

\bibitem[\protect\citename{Ratnaparkhi}1997]{rat-97}
Ratnaparkhi, A.,
\newblock 1997,
\newblock A linear observed time statistical parser based on maximum entropy
  models,
\newblock In {\em EMNLP2}, Providence, {RI}.

\bibitem[\protect\citename{Ron, Singer, and Tishby}1995]{rst-apfa-95}
Ron, D., Singer, Y., and Tishby, N.,
\newblock 1995,
\newblock On the learnability and usage of acyclic probabilistic finite
  automata,
\newblock In {\em Proceedings of the 8th Annual Conference on Computational
  Learning Theory ({COLT}'95)}, pp. 31--40, New York, NY, USA, ACM Press.

\bibitem[\protect\citename{Roth}1998a]{roth-snow}
Roth, D.,
\newblock 1998a,
\newblock Learning to resolve natural language ambiguities: A unified approach,
\newblock In {\em proc.~of the Fifteenth National Conference on Artificial
  Intelligence}, pp. 806--813, Menlo Park, CA, USA, AAAI Press.

\bibitem[\protect\citename{Roth}1998b]{roth-pc}
Roth, D.,
\newblock 1998b,
\newblock Private Communications.

\bibitem[\protect\citename{Satta and Henderson}1997]{sh-97}
Satta, G., and Henderson, J.~C.,
\newblock 1997,
\newblock String transformation learning,
\newblock In {\em Proc.~of the ACL/EACL Annual Meeting}, pp. 444--451, Madrid,
  Spain.

\bibitem[\protect\citename{Schiller}1996]{schiller-96}
Schiller, A.,
\newblock 1996,
\newblock Multilingual finite-state noun phrase extraction,
\newblock In {\em ECAI '96 Workshop on Extended Finite State Models of
  Language}, pp. 65--69, Budapest.

\bibitem[\protect\citename{Sekine}1998]{sekine-phd}
Sekine, S.,
\newblock 1998,
\newblock {\em Corpus-Based Parsing and Sublanguage Studies},
\newblock {Ph.D.} thesis, New York University.

\bibitem[\protect\citename{Sima`an}1996]{simaan-96}
Sima`an, K.,
\newblock 1996,
\newblock Computational complexity of probabilistic disambiguation by means of
  tree grammars,
\newblock In {\em Proceedings of the seventeenth International Conference on
  Computational Linguistics (COLING'96)}, pp. 1175--1180, Copenhagen, Denmark.

\bibitem[\protect\citename{Skut and Brants}1998]{skut-brants-98}
Skut, W., and Brants, T.,
\newblock 1998,
\newblock A maximum-entropy partial parser for unrestricted text,
\newblock In {\em Proc.~of the sixth Workshop on Very Large Corpora}, Montreal,
  Canada.

\bibitem[\protect\citename{Stanfill and Waltz}1986]{Stanfill-Waltz86}
Stanfill, C., and Waltz, D.,
\newblock 1986,
\newblock Toward memory-based reasoning,
\newblock {\em Communications of the ACM}, 29(12):1213--1228.

\bibitem[\protect\citename{van Rijsbergen}1979]{vanrijs-79}
van Rijsbergen, C.~J.,
\newblock 1979,
\newblock {\em Information Retrieval},
\newblock Buttersworth.

\bibitem[\protect\citename{Veenstra}1998]{veenstra-np-98}
Veenstra, J.,
\newblock 1998,
\newblock Fast np chunking using memory-based learning techniques,
\newblock In F.~Verdenius and W.~van~den Broek, editors, {\em Proceedings of
  Benelearn}, pp. 71--79, Wageningen, the Netherlands.

\bibitem[\protect\citename{Yvon}1996]{yvon-96}
Yvon, F.,
\newblock 1996,
\newblock Grapheme-to-phoneme conversion using multiple unbounded overlapping
  chunks,
\newblock In {\em Proceedings of the conference on New Methods in Natural
  Language Processing (NeMLaP II)}, pp. 218--228, Ankara, Turkey.

\end{thebibliography}

\end{document}